\documentclass{article}

    \usepackage[preprint]{neurips_2024}

\usepackage[utf8]{inputenc} % allow utf-8 input
\usepackage[T1]{fontenc}    % use 8-bit T1 fonts
\usepackage{hyperref}       % hyperlinks
\usepackage{url}            % simple URL typesetting
\usepackage{booktabs}       % professional-quality tables
\usepackage{amsfonts}       % blackboard math symbols
\usepackage{nicefrac}       % compact symbols for 1/2, etc.
\usepackage{microtype}      % microtypography
\usepackage{xcolor}         % colors
\usepackage{amsmath}
\usepackage{amssymb}
\usepackage{graphicx}
\usepackage{hyperref}
\usepackage{booktabs}
\usepackage{xcolor}
\usepackage{mdframed}
\usepackage{tcolorbox}
\usepackage{fontawesome}
\usepackage{geometry}
\usepackage{algorithm}
\usepackage{algpseudocode}
\usepackage{booktabs}
\usepackage{array}
\usepackage{caption}
\usepackage{amsmath, amssymb, amsthm}
\usepackage{thmtools}

\declaretheoremstyle[
    headfont=\normalfont\bfseries,
    bodyfont=\normalfont\itshape,
    spaceabove=6pt,
    spacebelow=6pt,
    headpunct={.},
]{mystyle}

\declaretheorem[style=mystyle,name=Definition]{definition}

\declaretheorem[style=mystyle,name=Theorem]{theorem}

\newmdenv[
  linecolor=gray!50,
  linewidth=0.5pt,
  topline=true,
  bottomline=true,
  leftline=true,
  rightline=true,
  backgroundcolor=gray!5,
  innerleftmargin=5pt,
  innerrightmargin=5pt,
  innertopmargin=5pt,
  innerbottommargin=5pt,
  skipabove=10pt,
  skipbelow=10pt
]{promptframe}

\definecolor{lightgray}{RGB}{240,240,240}

\newcommand{\keyterm}[1]{\textit{\textcolor{darkgray}{#1}}}

\newcommand{\lr}[1]{\left(#1\right)}

\title{Networks of Networks: Complexity Class Principles Applied to Compound AI Systems Design}

\author{%
  Jared Quincy~Davis\thanks{Corresponding author. Email address: jared@mlfoundry.com; jaredq@cs.stanford.edu} \\
  Foundry\\
  Stanford University\\
  \texttt{jared@mlfoundry.com} \\
  \And
  Boris Hanin \\
  Foundry \\
  Princeton University \\
  \And
  Lingjiao Chen \\
  Stanford University \\
  \And
  Peter Bailis \\
  Stanford University \\
  \And
  Ion Stoica \\
  University of California, Berkeley \\
  \And
  Matei Zaharia \\
  University of California, Berkeley \\
}

\begin{document}

\maketitle

\begin{abstract}
    As practitioners seek to surpass the current reliability and quality frontier of monolithic models, Compound AI Systems consisting of many language model inference calls are increasingly used. In this work, we construct systems, which we call Networks of Networks (NoNs) organized around the distinction between generating a proposed answer and verifying its correctness, a fundamental concept in complexity theory that we show empirically extends to Language Models (LMs). We introduce a verifier-based judge NoN with K generators, an instantiation of "best-of-K" or "judge-based" compound AI systems. Through experiments on synthetic tasks such as prime factorization, and core benchmarks such as the MMLU, we demonstrate notable performance gains. For instance, in factoring products of two 3-digit primes, a simple NoN improves accuracy from 3.7\% to 36.6\%. On MMLU, a verifier-based judge construction with only 3 generators boosts accuracy over individual GPT-4-Turbo calls by 2.8\%. Our analysis reveals that these gains are most pronounced in domains where verification is notably easier than generation--a characterization which we believe subsumes many reasoning and procedural knowledge tasks, but doesn't often hold for factual and declarative knowledge-based settings. For mathematical and formal logic reasoning-based subjects of MMLU, we observe a 5-8\% or higher gain, whilst no gain on others such as geography and religion. We provide key takeaways for ML practitioners, including the importance of considering verification complexity, the impact of witness format on verifiability, and a simple test to determine the potential benefit of this NoN approach for a given problem distribution. This work aims to inform future research and practice in the design of compound AI systems.
\end{abstract}

\section{Introduction}

In scenarios where practitioners are willing to expend a higher budget to go beyond the current reliability and quality frontier achievable via today's state-of-the-art monolithic models, Compound AI Systems consisting of many language model inference calls are increasingly employed ~\cite{compound-ai-blog, AlphaCode2T, stechly2023gpt, chen2024more, valmeekam2023can}. Implicitly or explicitly, we are often constructing networks of these calls: networks of networks, of sorts. This begs the question of what principles can guide the structure of the composition of these networks. As a central idea in this direction, we propose constructions organized around an explicit delineation between \textit{generation complexity} and \textit{verification complexity}. Correctly answering a question typically involves both \textit{generating} a proposed answer and \textit{verifying} that this answer is correct. This distinction between generation and verification is an organizing principle in complexity theory since, in many cases, the fastest algorithms for generation are more computationally expensive than generic algorithms for verification.

The key message of this article is that verification vs. generation complexity asymmetry notions continue to hold empirically when replacing tailored algorithms with Language Models (LMs). That this should be the case is not obvious, because it is far from clear what sorts of algorithms LMs use when asked to either generate or verify. We demonstrate our findings using a simple compound LM inference system that explicitly separates generation and verification calls. Specifically, we consider a compound AI system we term a \textit{verifier-based judge system with $K$ generators}, which works as follows:
\begin{itemize}
    \item A query/question $q$ is fed independently into $K$ language model calls, which can be different models or the same model $K$ times (we typically use GPT-4o or GPT4-turbo). $K$ is a hyper-parameter.
    \item For $k=1,\ldots, K$ the $k^{th}$ LM generator outputs $G_k(q) = (a_k,\pi_k)$, where $a_k$ is the proposed answer and $\pi_k$ is the proposed explanation for why this answer is correct. 
    \item Answer-explanation pairs $(a_k,\pi_k)$ are provided one at a time to another LM, which acts as a verifier or judge. For each $k=1,\ldots, K$, the LM returns $J(q,a_k,\pi_k)\in \{0,1\}$. An output of $1$ means $a_k$ correctly answers the question $q$, given the explanation $\pi_k$. If $J(q,a_k,\pi_k)=1$, the answer $a_k$ is accepted and the algorithm terminates and returns $a_k$. 
    \item If all answers are rejected, then a random answer is returned. 
\end{itemize}
Note that these verifier-based judges explicitly separate generation and verification. We find that they lead to notable performance gains on both synthetic and benchmark algorithmic reasoning tasks.  

For example, we consider factoring an integer into a product of two primes. This is a canonical problem for which it is widely believed that generation is computationally expensive since it requires factoring, while verification is fast because it only involves multiplication. We find that individual calls to GPT-4o have a very low probability of correctly factoring even numbers $p$ which are the product of 2 three digit integers (around $3.7\%$), while the same LM has approximately a $90\%$ probability of correctly deciding whether a given factorization is valid. In particular, using a judge-based verifier system with $K=10$ generators allows us to improve the performance from $3.7\%$ to $36.6\%$ (see Table \ref{tab:factorization_results}). This matches almost exactly the probability of at least one generator proposing a correct factorization, as predicted by the our analysis in \S \ref{sec:theory}.

This simple example suggests that when using compound systems made up of many calls to an LM it can be useful in problems with hard generation and relatively easy verification to invest more resources into generation than verification. This is precisely what is done in invoking chain of thought with Gemini \cite{team2023gemini} and in AlphaCode2 \cite{li2022competition}.

To further validate this beyond factorization tasks, we study the performance of verifier-based judge systems on MMLU \cite{hendrycks2020measuring}. We continue to find significant improvements in overall performance. Individual GPT-4-Turbo calls obtain an accuracy of $81.2\%$, but a verifier-based judge with $K=3$ generators obtains $84\%$ accuracy. The gains are magnified when broken down by topic, with subjects such as Abstract Algebra, Formal Logic, High School Math, and High School Physics--which arguably exhibit especially large gaps between the complexity of generation and verification--showing gains ranging from $6.35\%$ to $8.76\%$ (see Tables \ref{tab:accuracy_gain} and \ref{tab:accuracy_gain2}). 

Based upon our results, we encourage ML practitioners to \textbf{Think about verification complexity}. Some tasks will be more verifiable than others. Most traditional computing and software engineering tasks, in particular, possess this property. Our results suggest this asymmetry extends to LMs as well, when prompted to act as verifiers; however, you can also employ classical simulators, unit tests, or other components as verifiers in Compound AI Systems design. Also, note that given questions sampled from a distribution, \textbf{there is a simple test} for (i) whether the approach proposed in this paper will help for subsequent problems, as expressed in \S \ref{sec:theory}, and (ii) to what extent.

We hope practitioners will find these discussions helpful in informing future research and practice in this emerging domain.

\section{Verifier-Based Judges: Formalism and Analysis}

\subsection{Construction}

In this section, we introduce verifier-based judges. 

\begin{definition}[Verifier-Based Judge]
A verifier-based judge $J$ for a query $q$ and $K$ answer-witness pairs $(\mathbf{a}, \mathbf{\pi}) = ((a_1, \pi_1), \ldots, (a_K, \pi_K))$ operates as follows:

Let $\sigma$ be a random permutation of $\{1, \ldots, K\}$. Then:

\begin{equation}
J(q, (\mathbf{a}, \mathbf{\pi})) = 
\begin{cases}
a_{\sigma(i)} & \begin{array}{@{}l@{}}
    \text{if } \exists i \leq K : V(q, a_{\sigma(i)}, \pi_{\sigma(i)}) = 1 \\
    \text{and } \forall j < i, V(q, a_{\sigma(j)}, \pi_{\sigma(j)}) = 0
\end{array} \\[10pt]
\text{random } a_i & \text{if } \forall i \leq K, V(q, a_{\sigma(i)}, \pi_{\sigma(i)}) = 0
\end{cases}
\end{equation}
\end{definition}

Note that this is a version-zero construction without notions of verifier confidences. If no answer is accepted by the verifier, a random answer is returned. This is described in pseudo-code in Algorithm 1.

\begin{algorithm}[t]\label{alg:VBJ}
\caption{Verifier-Based Judge}
\begin{algorithmic}[1]
\Require Query $q$, Answer-witness pairs $(a_1, \pi_1), \ldots, (a_K, \pi_K)$, Verifier $J$
\Ensure Selected answer $a$ or random answer
\State Shuffle $(a_1, \pi_1), \ldots, (a_K, \pi_K)$ randomly
\For{$i \gets 1$ to $K$}
    \If{$J(q, a_i, \pi_i) = 1$}
        \State \Return $a_i$
    \EndIf
\EndFor
\State \Return Random answer from $\{a_1, \ldots, a_K\}$
\end{algorithmic}
\end{algorithm}

Intuitively, a single verifier-based judge operating over many generators should be effective when: (i) each generator individually exhibits moderate variance; (ii) different generators are not too correlated; (iii) for a typical query, the probability a typical generator gives the correct answer is low; (iv)  it is notably ~\textit{easier} for the verifier to discern the correct answer-witness pair than it is for the generators to produce it. In other words, ~\textbf{verification is easier than generation}. Formally, this final statement corresponds to requiring for a correct answer-witness pair $(a^*, \pi^*)$ that
    \[ P(V(q, a^*, \pi^*) = 1) > P(G_i(q) = (a^*, \pi^*)) \text{ for any generator } G_i .\]
We make this precise in the following section. 

\subsection{Verifier-based Judge Performance Characterization}\label{sec:theory}

In this section, we determine analytically the performance of a verifier-based judge on a given query $q$ with a unique correct answer $a^*$ in the simple case where we have access to $K$ \textit{iid} generators. Note that this iid assumption won't strictly hold in practice in many cases, but empirical results concord with this analysis well for the instance when generators are obtained by sampling with a fixed high temperature and independent random seeds from the outputs of a given LM\footnote{Future work can explore method to increase the observed diversity.}. The performance of the verifier-based judge across a collection of queries is then simply obtained by averaging the results in this section over the query instance. We express our answers in terms of the three fundamental quantities:
\begin{align*}
    c &:= \mathbb P\lr{J(q, (a,\pi))=1~|~a = a^*}\\
    s&:= \mathbb P\lr{J(q, (a,\pi))=0~|~a \neq a^*}\\
    r(q) &:= \mathbb P(G(q) = (a^*,\pi) \text{ for some }\pi).
\end{align*}
These quantities are the analogs of classical PCP notions of completeness, soundness, and generation complexity that are ubiquitous in complexity theory (see e.g. \cite{10.1145/273865.273901}).

\begin{theorem}[Verifier-Based Judge Performance]
Consider a verifier-based judge system with $K$ iid generators, and fix a query $q$ with a unique correct answer $a^*$. We have
\begin{equation}\label{eq:p-correct}
\mathbb P(J(q) = a^*) = rc\frac{1-(1-\beta)^K}{\beta} + (1-\beta)^K\frac{1}{|\mathcal A|},\qquad \beta:=1-((1-c)r + s(1-r)).
\end{equation}
\end{theorem}
\begin{proof}
    We have
    \begin{align*}
        \mathbb P\lr{\text{verifier correct}}&=\sum_{j=1}^K \mathbb P\lr{G_j\text{ correct and accepted},\, G_i\text{ rejected }\forall\, i<j}\\
        &+\mathbb P\lr{G_1,\ldots, G_K\text{ rejected, random answer correct}}\\
        &=\sum_{j=1}^K \mathbb P\lr{G_j\text{ correct}}\mathbb P\lr{G_j \text{ accepted}~|~G_j\text{  correct}} \prod_{i<j}\mathbb P\lr{G_i\text{ rejected}}\\
        &+\prod_{i<j}\mathbb P\lr{G_i\text{ rejected}}\frac{1}{|\mathcal A|},
    \end{align*}
    where for the second equality we have used the independence between both generators and the verifier. Note that, by definition, 
    \[
\mathbb P\lr{G_j\text{ correct}}\mathbb P\lr{G_j \text{ accepted}~|~G_j\text{  correct}}  = rc,\qquad \mathbb P\lr{G_i\text{ rejected}} = \beta. 
    \]
    Summing on $j=1,\ldots, K$ completes the derivation. 
\end{proof}

To complete our analysis in this section, we seek to describe how much we expect to gain with a verifier-based judge above the natural baseline of using a single generator by computing the difference
\[
G(q, K) = P(J(q) = a^*) - r(q)
\]
between the probability that a verifier-based judge returns the correct answer and the probability that a single generator returns the correct answer. 

\begin{theorem}[Verifier-Based Judge Gain]
When the number of generators $K$ goes to infinity, the gain of a verifier-based judge is positive if and only if:
\begin{equation}
r\neq 0,1\qquad \text{and}\qquad c=\mathbb P\lr{J(a,\pi)=1~|~a=a^*} > \mathbb P\lr{J(a,\pi)=1~|~a\neq a^*}=1-s.
\end{equation}
The second condition is precisely the statement that the judge is more likely to accept a correct answer than an incorrect answer.
\end{theorem}
\begin{proof}
    Taking $K\rightarrow \infty$ in \eqref{eq:p-correct} gives
    \[
\lim_{K\rightarrow \infty}\bigg[\mathbb P\lr{J(q)=a^*}-r\bigg] = r\lr{\frac{c}{\beta}-1}.
    \]
    Note that
    \begin{align*}
c &= \mathbb P\lr{J(a,\pi)=1~|~a=a^*}\\
\beta &=\mathbb P\lr{J(a,\pi)=1}\\
&=  \mathbb P\lr{J(a,\pi)=1~|~a=a^*}\mathbb P\lr{a=a^*}+\mathbb P\lr{J(a,\pi)=1~|~a\neq a^*}\mathbb P\lr{a\neq a^*}.
    \end{align*}
Hence, 
\[
c> \beta\qquad \Longleftrightarrow\qquad \mathbb P\lr{J(a,\pi)=1~|~a=a^*} > \mathbb P\lr{J(a,\pi)=1~|~a\neq a^*},
\]
as desired.
\end{proof}

This theorem shows that verifier-based judge system confers a gain, at least with a large number of generators, as soon as it's completeness exceeds a threshold determined by its soundness. Moreover, the extent of this gain is proportional to the query difficulty. 

Our analysis also reveals the diminishing returns of increasing the ensemble size $K$. While larger ensembles generally improve performance, the marginal benefit decreases, suggesting an optimal trade-off point between accuracy and computational cost. We plan to probe this more extensively in future work. 

These insights can guide the design and optimization of compound AI systems. For instance, they suggest that resources might be better spent improving verifier quality or invoking the verifier repeatedly, rather than increasing ensemble size beyond a certain point. They also illustrate that such systems can be particularly valuable for difficult queries where individual generators are unreliable, provided that verification is sufficiently easier than generation for the given domain.

\section{Experiments}

In this section we present the two kinds of experiments discussed in the Introduction. First, in \S \ref{sec:factorization}, we detail experiments using LMs for prime factorization. Then, in \S \ref{sec:mmlu}, we break down our experiments on MMLU. 

\subsection{Prime Factorization}
\label{sec:factorization}

As an example problem to demonstrate that sometimes verification is easier than generation for LMs, we invoke and probe the classic problem of ~\keyterm{prime factorization}.

The prime factorization problem is core to cryptography and foundational algorithms like RSA ~\citep{10.1145/359340.359342}. It turns out it is very laborious to factor a number that is a product of two large primes into the constituent primes; however, it is relatively easy to multiply two primes. Thus, if I have two candidate constituent primes, it can be easy and relatively quick to multiply them to test if they constitute a valid factorization.

Building on this intuition, we conjectured that a simple verifier--based judge system could outperform a baseline of single LM calls. To understand this, we first implemented two components:

\begin{itemize}
    \item \textbf{Generators} tasked with factorization.
    \item \textbf{Verifiers} tasked with assessing a factorization and deeming it correct or incorrect.
\end{itemize}

We started off using GPT-4o and asking it to help us factor numbers that are a product of two 3-digit primes. This yielded a \textbf{3.70\% success rate} for the generators. For the verifier task of assessing a factorization as correct or incorrect, a GPT-4o-based verifier was able to correctly classify a proposed factorization \textbf{90.11\% of the time}. Moreover, these results were well calibrated as excited by the completeness and soundness of the verifier, which intuitively correspond to the true positive and true negative rates. 

These initial results convinced us to compose these components into a verifier-based judge system. Based on those results, the baseline GPT-4o in this case model achieved the \textbf{3.7\%} on up-to-3 digit factorization and a verifier-based judge system with $K=10$ generators achieved \textbf{36.6\%}\footnote{We produced these numbers using $10,000$ synthetically generated examples.}.

\begin{table}[ht]
\centering
\renewcommand{\arraystretch}{1.5}
\begin{tabular}{@{}lccc@{}}
\toprule
\textbf{Task} & \textbf{GPT-4o (\%)} & \textbf{VBJ, K=10, GPT-4o (\%)} & \textbf{$\Delta$ (\%)} \\
\midrule
3-digit Factorization & 3.7 & \textbf{36.6} & 29.9 \\
\midrule
\multicolumn{4}{l}{\hspace{-.22cm}\textbf{Component Performance:}} \\
 Generator task accuracy  (\%) & 3.7 & &  \\
 Verification task accuracy (\%), c, s & \textbf{90.1} & .97 & .9 \\
\bottomrule
\end{tabular}
\vspace{10pt}
\caption{Factorization results comparing the baseline GPT-4o model with a 10 Ensemble, 1 Verifier-Based Judge system using GPT-4o also. The component performance of the Generator and Verifier (both using GPT-4o) in isolation is also shown. A verifier is able to verify a candidate answer as correct or incorrect with high accuracy. Thus, when given a correct candidate answer, it has a high chance of selecting the correct answer. The $36.6\%$ accuracy  is notably congruent with the theoretical predictions for iid generations, since we found that the empirical estimates for completeness and soundness where $.97$ and $.9$, respectively.}
\label{tab:factorization_results}
\end{table}

\subsection{Lottery Ticket Problem}
As another sub-experiment, we looked at a problem we viewed as classically non-verifiable, which we call the \keyterm{lottery ticket problem}. In this problem, we have an oracle that picks a number within some range (0 and 100, for example). The task of the generator models is to guess the number. We then give the N guesses from the generators to the judge model and asks it to, optionally considering the guesses from the generators ("advisors") guess the oracle's number. As one might expect, a verifier-based judge confers no advantage in this case and success is roughly $1/N$ regardless.  

\subsection{MMLU}\label{sec:mmlu}

Beyond these specific examples, we also ran probes on the Massive Multitask Language Understanding (MMLU) benchmark ~\citep{hendrycks2020measuring}.

On the MMLU, we ran a baseline model of just gpt-4-1106-preview with temperature $1.0$. We explored a number of temperatures in early probing and observed no clear performance gain pattern. What we did notice, though, was that our baseline models all seemed to under-perform the official reported numbers from the GPT-4 technical report. This result, though, is congruent with other published work, such as ~\cite{zheng2023gpt} which claimed a 78.3\% 5-shot performance of GPT-4-Turbo on MMLU. We also note that the Official OpenAI GPT-4 Technical reported results using the \textbf{base model}. OpenAI, in the GPT-4 Technical Report \citep{achiam2023gpt} states that the post-training, alignment, and instruction fine-tuning, which give rise to the friendly and helpful assistant persona we interact with, "meaningfully hurts calibration" on some of these concrete reasoning tasks and benchmarks, and specifically highlights MMLU. 

Running GPT-4-Turbo vs. the verifier-based judge system with $K=3$ generators, we see that even this basic compound system beats the baseline GPT-4-Turbo model by a 2.8\% margin in accuracy. To put this gain in perspective, the difference between MMLU scores for models like Gemini 1.5 ~\cite{team2023gemini}, Llama-400b ~\cite{Llama3-blog} (early snapshot), and Claude 3 ~\cite{claude3-blog} is within 1\%. 

\begin{table}[ht]
\centering
\renewcommand{\arraystretch}{1.5} % Increase the row height
\begin{tabular}{@{}lccc@{}} % Add padding within cells
\toprule
\textbf{} & \textbf{GPT-4-Turbo (\%)} & \textbf{VBJ, K=3, GPT-4-Turbo (\%)} & \textbf{$\Delta$ (\%)} \\
\midrule
MMLU & 81.2 & \textbf{84.0} & +2.80 \\
\bottomrule
\end{tabular}
\vspace{10pt} % Add vertical space between the table and the caption
\caption{Overall MMLU Results. A simple verifier-based judge system with $K=3$ beats the baseline GPT-4 in performance at a tradeoff of higher latency and cost.}
\label{tab:accuracy_gain}
\end{table}

% \subsubsection{Varied gain across domains}

While the overall gain from 81.2\%\footnote{0.17 stddev over 3 full runs} to 84.0\%\footnote{0.14 stddev over 3 full runs} is intriguing, even more noteworthy is that the gain is far from uniformly distributed. There are some subjects for which the verifier-based judges systems confer much more substantial gain, and others for which it seems to make no noticeable or statistically significant difference. 

The subjects for which the verifier-based judges systems seemed to confer the greatest advantage are listed in Table \ref{tab:accuracy_gain}. Similarly, the subjects with lowest gains are listed in Table \ref{tab:accuracy_gain2}. The latter, we note, seem to be subjects where factual and declarative knowledge is most critical. 

\begin{table}[ht]
\centering
\renewcommand{\arraystretch}{1.5} % Increase the row height
\begin{tabular}{@{}lccc@{}} % Add padding within cells
\toprule
\textbf{Subject} & \textbf{GPT-4-Turbo (\%)} & \textbf{VBJ, K=3, GPT-4-Turbo (\%)} & \textbf{$\Delta$ (\%)} \\
\midrule
Abstract Algebra & 72.05 & 80.81 & 8.76 \\
Formal Logic & 68.25 & 76.98 & 8.73 \\
High School Math & 78.56 & 86.99 & 8.43 \\
High School Physics & 77.92 & 84.77 & 6.85 \\
Electrical Engineering & 72.64 & 78.62 & 5.98 \\
Professional Accounting & 81.44 & 86.88 & 5.44 \\
College Mathematics & 65.73 & 71.03 & 5.30 \\
\bottomrule
\end{tabular}
\vspace{10pt} % Add vertical space between the table and the caption
\caption{Verifier-based judge system with $K=3$ generators performance vs. the single GPT-4-Turbo baseline performance on MMLU. The subjects where the verifier-based judge provides the greatest gain seem to be domains where procedural reasoning is more critical than declarative/factual knowledge.}
\label{tab:accuracy_gain}
\end{table}

\begin{table}[ht]
\centering
\renewcommand{\arraystretch}{1.5} % Increase the row height
\begin{tabular}{@{}lccc@{}} % Add padding within cells
\toprule
\textbf{Subject} & \textbf{GPT-4-Turbo (\%)} & \textbf{VBJ, K=3 GPT-4-Turbo (\%)} & \textbf{$\Delta$ (\%)} \\
\midrule
High School Psychology & 94.37 & 94.86 & 0.49 \\
College Chemistry & 65.67 & 66.00 & 0.33 \\
World Religions & 86.35 & 86.55 & 0.19 \\
High School Geography & 93.27 & 92.93 & -0.34 \\
\bottomrule
\end{tabular}
\vspace{10pt} % Add vertical space between the table and the caption
\caption{Verifier-based judge system with $K=3$ generators vs. the single GPT-4-Turbo baseline performance on MMLU. The subjects where the verifier-based judge provides the least gain seem to be domains where declarative/factual knowledge is more critical than multi-step procedural reasoning.}
\label{tab:accuracy_gain2}
\end{table}

To better understand how verifier-based judges help across query difficulty, we present in Table \ref{tab:performance_metrics} sub-results for the electrical engineering subject in MMLU. The verifier-based system, using an ensemble of generators and a single judge, achieved an aggregate accuracy of 78.62\%, outperforming the individual generator accuracy of 72.64\%. The breakdown of performance improvements shows that the verifier-based system help in questions for which the individual generators had divergent responses (only 1 or 2 out of 3 generators agreed on the correct answer), since the judge was able to discern the correct answers with higher likelihood than the generators could produce them.

These results demonstrate how a verifier-based judges can confer a gain if the judge construction amplifies the base accuracy of the generators.

\begin{table}[ht]
\centering
\renewcommand{\arraystretch}{1.2}
\begin{tabular}{@{}lcccc@{}}
\toprule
\multicolumn{5}{c}{\textbf{Subject: Electrical Engineering (N = 145)}} \\
\multicolumn{2}{l}{\textbf{Ver.-Bas. Jug. aggregate accuracy:} 78.62\%} & \multicolumn{3}{r}{\textbf{ Generator accuracy:} 72.64\%} \\
\midrule
\textbf{Gen. Acc. (\%)} & \textbf{Ver.-Bas. Jug. (\%)} & \textbf{\% in Group} & \textbf{Improvement (\%)} \\
\midrule
100.0 & 100.0 & 60.7 & 0.0 \\
66.7 & 94.1 & 11.7 & 27.5 \\
33.3 & 55.6 & 12.4 & 22.2 \\
0.00 & 0.0 & 15.2 & 0.0 \\
\bottomrule
\end{tabular}
\vspace{10pt} % Add vertical space between the table and the caption
\caption{Performance breakdown of a verifier-based judge system with $K=3$ generators for the electrical engineering subject in MMLU. For 11.7\% and 12.4\% of queries, 2 or 1 out of 3 generators got the answer correct, but the verifier-based judge was able to discover the correct answer in 94.1\% and 55.6\% of those cases, respectively.}
\label{tab:performance_metrics}
\end{table}

\section{Related Work}

The concept of composing multiple calls to models to improve performance has been implicitly invoked repeatedly in recent months. The Gemini Technical Report employed a novel scheme at inference time, termed CoT@32, as opposed to the default 5-shot prompting invocation scheme ~\cite{team2023gemini}. DeepMind's AlphaCode 2, in order to go "beyond the capabilities of existing AI systems," employed a method that uses large-scale transformers to generate "up to a million code samples per problem" and then filters down the set to converge to a solution. ~\citet{li2022competition}. N-shot prompting itself implicitly is a systems approach in that it often elicits longer (and thus higher latency / costlier) responses from models but can result in improved performance. ~\cite{wei2022chain}. ~\citet{chen2024more, li2024more} probed how calling an LLM many times can improve or degrade performance on a task set, depending on the distribution of query difficulty in the set. Numerous other recent works have explored iteratively chaining self-improvement calls ~\cite{madaan2024self, wang2024mixture, valmeekam2023can, stechly2023gpt},

These self-improvement or ensemble systems can perhaps be unified and viewed through a single framework as wide or deep networks (or meshes) of calls to a given model or a cross a set of models and modules. 

\paragraph{Self-Verification and Collections of Model Calls}

\citet{stechly2023gpt, valmeekam2023can} called into question claims that models can self-critique and iteratively self-refine ~\citep{madaan2024self} their solutions when invoked in deep chains of calls. ~\citet{stechly2023gpt} evaluated iterative self-improvement claims over a suite of graph coloring problems and found that GPT4 achieves sub 20\% accuracy; however, perhaps most surprisingly, they found the accuracy diminished in "self-critiquing mode." Their work explained this by noting that even when GPT4 can guess a valid coloring by chance, its self-critiquing might lead to degradation in performance. This is congruent with results in ~\citet{chen2024more}, which notes that making more calls to what it called voting-based LLM systems (ensembles with a voting-based aggregation function) can diminish performance for "hard" queries due to a sort of variance-reduction that hurts questions where the likelihood of a correct answer is <50\% since the model might otherwise have "gotten lucky". Conversely, other recent works like ~\cite{li2024more, madaan2024self} also explored calling models in structured forms and demonstrated that this could lead to performance gain, even showing that systems consisting of weak LLMs can surpass the performance of the best individual LLM in social conversation settings~\cite{wang2024mixture}. 

These seemingly disparate results can be reconciled, perhaps. ~\citet{stechly2023gpt, valmeekam2023can}, suggest that when an "externally sound verifier" module is leveraged, "try again" systems methods can induce performance gains. This suggests that networks of calls can engender gains, but only under certain conditions. 

We hope that our study and the discussion of the concept of verifiability complexity can shine a light on some of those conditions.

\paragraph{Classical complexity notions of verification and generation.} Comparison of verification and generation has been extensively studied via the lens of the classic computational complexity~\cite{papadimitriou2003computational,courcoubetis1995complexity,clarke1997model}. In fact, one of the most popular classes of problems, NP problems, are those whose solutions can be verified in polynomial time. Pioneering theoretical computer science (TCS) work~\cite{courcoubetis1995complexity} places deep emphasis on understanding how verification can be leveraged to provably generate (approximately) high-quality solutions to hard problems. In the context of LMs, however, it is unclear which ``algorithm'' an LLM uses and thus which ``complexity'' it incurs to solve a problem. This paper, though, demonstrates that the notion of verification and generation complexity can still apply to LMs, and can be leveraged to inform the design of compound LLM systems.

\section{Conclusion and Future Work}

Here we motivate the consideration of complexity class principles in Compound AI Systems design. We argue that LM practitioners should consider, measure, and build around the potential verification vs. generation difficulty asymmetry in their domain of interest. 

In subsequent work, we hope to more extensively probe the verifier-based compound AI systems architecture landscape, investigating questions around how to \textbf{measure and maximize the diversity among generator outputs}, how the \textbf{verifier's ability to discriminate varies} as a function of tunable parameters, and more. We also hope to release code that the community can build upon to contribute to this further probe this theme.

\bibliographystyle{plainnat}
\bibliography{references}

\end{document}